\documentclass[sigconf]{acmart}

\copyrightyear{2019} 
\acmYear{2019} 
\setcopyright{acmlicensed}
\acmConference[ICAIL '19]{Seventeenth International Conference on Artificial Intelligence and Law}{June 17--21, 2019}{Montreal, QC, Canada}
\acmBooktitle{Seventeenth International Conference on Artificial Intelligence and Law (ICAIL '19), June 17--21, 2019, Montreal, QC, Canada}
\acmPrice{15.00}
\acmDOI{10.1145/3322640.3326740}
\acmISBN{978-1-4503-6754-7/19/06}

\begin{document}
\title[Legal Case Retrieval via Lexical Matching and Summarization]{Building Legal Case Retrieval Systems with Lexical Matching and Summarization using A Pre-Trained Phrase Scoring Model}

\author{Vu Tran}
\email{vu.tran@jaist.ac.jp}
\affiliation{
  \institution{Japan Advanced Institute of Science and Technology}
}

\author{Minh Le Nguyen}
\email{nguyenml@jaist.ac.jp}
\affiliation{
  \institution{Japan Advanced Institute of Science and Technology}
}

\author{Ken Satoh}
\email{ksatoh@nii.ac.jp}
\affiliation{
  \institution{National Institute of Informatics}
  \country{Japan}}

\renewcommand{\shortauthors}{Tran et al.}

\begin{abstract}
We present our method for tackling the legal case retrieval task of the Competition on Legal Information Extraction/Entailment 2019. Our approach is based on the idea that summarization is important for retrieval. On one hand, we adopt a summarization based model called encoded summarization which encodes a given document into continuous vector space which embeds the summary properties of the document. 
We utilize the resource of COLIEE 2018 on which we train the document representation model. On the other hand, we extract lexical features on different parts of a given query and its candidates. We observe that by comparing different parts of the query and its candidates, we can achieve better performance. Furthermore, the combination of the lexical features with latent features by the summarization-based method achieves even better performance. 
We have achieved the state-of-the-art result for the task on the benchmark of the competition.

\end{abstract}




\maketitle

\section{Introduction}
Automatic legal document processing systems can speed up significantly the work of experts, which, otherwise, requires significant time and efforts. One crucial kind of such systems, automatic information retrieval whose systems, in place of experts, process over enormous amount of documents, for example, legal case reports and statute law documents, which are accumulated rapidly over time. One of the challenging tasks is legal case retrieval task, where the corresponding systems, in place of experts, process over enormous amount of legal case documents which are accumulated rapidly over time (the number of filings in the U.S. district courts for civil cases and criminal defendants is 344,787 in 2017 \footnote{http://www.uscourts.gov/statistics-reports/judicial-business-2017}). 

Legal case retrieval or retrieval of prior cases is an important research topic for decades where approaches to solve the corresponding task involve performing linguistics analysis, logical analysis, common lexical matching, and distributed vector representation with both common and legal expertise knowledge\cite{bench2012history}. In \cite{jackson2003information}, they build a system called ``History Assistant"  which extracts rulings from court opinions and retrieves relevant prior cases from a citator database by combining partial parsing techniques with domain knowledge and discourse analysis to extract information from the free text of court opinions.
In \cite{10.1007/11552413_49}, they develop a knowledge representation model for the intelligent retrieval of legal cases involving decomposing issues into sub-issues, and categorizing factors into pro-claimant, pro-responder and neutral factors. In \cite{Saravanan2009}, they overcome the problem of keyword-based search due to synonymy and ambivalence of words by developing an ontological framework to enhance the user’s query and ensure efficient retrieval by enabling inferences based on domain knowledge. Other works related to building legal ontology are \cite{wyner2008ontology,wyner2012legal,getman2014crowdsourcing}. Aside of linguistics approaches which are expensive to develop because of the required expertise knowledge, other approaches utilizes the emerging effectiveness of neural networks for natural language processing with the pioneer method of mapping texts to continuous vector space\cite{mikolov2013distributed,le2014distributed}. In \cite{mandal2017measuring}, the authors measure legal document similarity considering structural information of the document including paragraphs, summary and utilizing various representation methods including lexical features: TF-IDF, and topic modeling, and distributed vector representational features: \textit{word2vec}, and \textit{doc2vec}. 

The Competition on Legal Information Extraction/Entailment, for the first time in 2018, organized a legal case retrieval task~\cite{COLIEE2018}. The task was tackled in various approaches in both lexical matching and deep learning methods. The authors of the best system~\cite{vu2018coliee} have tackled the legal case retrieval task of COLIEE 2018 by developing a document encoding method using expert summary for training a phrase scoring model utilizing deep neural networks. In COLIEE 2019, we would like to utilize the resource of COLIEE 2018 and adopt the phrase scoring model pre-trained on COLIEE 2018 dataset to COLIEE 2019 dataset, and experiment the complementary benefits of using the encoding method with lexical matching for better retrieval system. 

In this paper, we describe our method for tackling the legal case retrieval task in Competition on Legal Information Extraction/Entailment (COLIEE), 2019 with the following main points:

\begin{itemize}
    \item We showed the effectiveness of representing documents in continuous vector space in which the summary properties of the documents are embedded. The representation method is based on a phrase scoring model which is trained to  assign high scores for phrases representing contexts that are similar to contexts found in the summary. 
    \item The method is used to generate not only the latent features but also surface texts as predicted summary for lexical matching.
    \item We utilized the resource of COLIEE 2018 on which we train the document representation model.
    \item We extracted lexical features on different parts of queries and candidates.
    \item We have achieved the state-of-the-art result for the task on the benchmark of the competition.
\end{itemize}

\section{Method}
\subsection{Overview of Legal Case Retrieval Task}
The legal case retrieval task involves reading a new case $q$, and extracting supporting cases $c^*_1$, $c^*_2$, ..., $c^*_n$ for the decision of $q$ from a given list of candidate cases. The candidate cases that support for the decision of a new case are called 'noticed cases'. The legal cases are sampled from a database of predominantly Federal Court of Canada case laws, provided by Compass Law.

In COLIEE 2018, when dealing with this task, Tran et al.~\cite{vu2018coliee} observed several obstacles. First, the candidate cases are $\approx$5.7K-token long in average (Table~\ref{tbl:data-stat-2018}). This poses the problem of understanding the reason of selecting the cases as supporting cases. They, then, chose another approach which is comparing the summaries of each query and its candidate cases. They, however, found that the summary of the query is not necessarily lexically similar to the summary of the candidate cases. Moreover, some candidate cases do not have summary at all. They obtained the summary for each and every candidate cases, and furthermore, the summary should be comparable with the summary of the corresponding query.
They came up with the idea of encoding the entire document into vector space embedding the properties of summarization, and called it encoded summarization. They realized the approach successfully for COLIEE 2018, and achieved the state of the art.

\begin{table}[H]
    \centering
    \caption{Statistics of candidate case documents in COLIEE 2018 training data.}
    \begin{tabular}{l|r|r}
    \hline \hline
        \textbf{Property}  & \textbf{Max} & \textbf{Avg.} \\
        \hline \hline
        \#words/doc & 85,551 & 5,690 \\
        \#paragraphs/doc & 1,117 & 43 \\
        \#summary-words/doc & 8,827 & 589 \\
        \hline \hline
    \end{tabular}
    \label{tbl:data-stat-2018}
\end{table}

\begin{table}[H]
    \centering
    \caption{Statistics of candidate case documents in COLIEE 2019 training data. (*) Only count documents having an expert summary.}
    \begin{tabular}{l|r|r}
    \hline \hline
        \textbf{Property}  & \textbf{Max} & \textbf{Avg.} \\
        \hline \hline
        \#words/doc & 9,666 & 2,665 \\
        \#paragraphs/doc & 119 & 22 \\
        \#summary-words/doc* & 3,085 & 242 \\
        \hline \hline
    \end{tabular}
    \label{tbl:data-stat-2019}
\end{table}
In COLIEE 2019, we observed the similar and different challenges. First, the candidate cases are $\approx$2.7K-token long in average (Table~\ref{tbl:data-stat-2019}). The difficulty of reading too long texts still emerges. We may pursue the idea that using summary as the main source of information.
However, the dataset of COLIEE 2019 is different from the one of COLIEE 2018. While in COLIEE 2018, most of the candidate cases have a summary, in COLIEE 2019, more than $\approx$47K in a total of 57K candidate cases are confirmed to have no summary (indicated with the note "This case is unedited, therefore contains no summary"). This means that summarization over candidate case requires additional effort so that we can compare a query's summary with a candidate's summary.

\subsection{Encoded Summarization} \label{sec:enc-summarization}
The summary of a document contains the highlights of the document, which is an important factor in relevance analysis. The summary, however, is short and may not cover enough information. 
In this section, we describe a process of summarization for both the query and the candidates to include more points from the whole document. 
For this purpose, we train a phrase scoring model for identifying important phrases which discuss contexts similar to those in the summary. 

In \cite{vutran2018mirel}, the authors present a way of obtaining catchphrases of legal case documents via phrase scoring using deep neural networks. Catchphrases represent important legal points of a legal case document and are usually drafted by experts, thus play important roles for understanding the case. To generate catchphrases, they built a scoring model to estimate phrasal scores of a legal case document and used the phrasal scores to construct predicted catchphrases for a new legal case. The scoring model is trained given expert drafted catchphrases of the document and optimized with the objective that catchphrases are expected to have higher scores than other contents in the document. In other words, we can train such a scoring model given training documents with indicated important contents which are document summaries in our task.
Inspired from the work's learning process, Tran et al.~\cite{vu2018coliee} proposed to obtain document embedding directed by document summary, applied successfully the method for COLIEE 2018, and achieved the state of the art.

The method, encoded summarization, is based on a phrase scoring model which is trained to assign high scores for phrases representing contexts that are similar to contexts found in the summary. 
Given a document, its the phrasal scores are estimated given its summary (extracted using indicators  `\textit{Summary:}' and `\textit{Present:}') and paragraphs. The phrase scoring model is trained with the objective that summary contents are expected to have higher scores than paragraph contents. We trained the model on COLIEE 2018 dataset only in which most of the candidate cases have an expert summary. Fine-tuning of the phrase scoring model over COLIEE 2019 dataset is left for future work. After obtaining the trained model, the final document representation in continuous vector space is composed from latent representations of the model. The encoded summary represents the summarization nature of the document. From the phrase scoring model, we also follow the phrase extraction in~\cite{vutran2018mirel} to generate text summary for lexical matching with summary of the query since most of the candidate cases in COLIEE 2019 dataset do not have a summary. 

\subsubsection{Phrase Scoring Model}
We describe the phrase scoring model of \cite{vutran2018mirel} for this task.

Given a document, we denote $w^{s_i}_{j}$ as word $j^{th}$ of  sentence $i^{th}$. The features of an n-gram phrase $p_{j}=\{w_{j}, w_{j+1}, ... ,w_{j+l-1}\}$ of a sentence are captured using convolutional neural layer as follows:  

\begin{equation}
\mathbf{f_{p_{j}}}=ReLU\left(\mathbf{W}^c  \left[ \begin{array}{l}
\mathbf{v}(w_{{j}}) \\
\mathbf{v}(w_{j+1}) \\
... \\
\mathbf{v}(w_{{j+l-1}}) \\

\end{array} 
\right]\right)
\end{equation}
where, 
$\mathbf{v}(\cdot):\ \mapsto \mathbb{R}^d$: word embedding vector lookup map,
$l$: corresponding to the window size containing $l$ contiguous words,
$[\cdot] \in \mathbb{R}^{dl}$: concatenated embedding vector,
$\mathbf{W}^c\in \mathbb{R}^{c\times dl}$: convolution kernel matrix with $c$ filters,
$\mathbf{f}_{p_{j}} \in \mathbb{R}^{c} $: phrase feature vector,
$ReLU$: rectified linear unit activation. 

Convolutional neural layers or convolutional neural networks have been applied widely in natural language processing, for example, text classification~\cite{kim:2014:EMNLP2014,kalchbrenner-grefenstette-blunsom:2014:P14-1,johnson-zhang:2015:NAACL-HLT}, machine translation~\cite{gehring2017convolutional}, text pair modeling (question answering~\cite{severyn2015learning}, and textual entailment~\cite{mou2016natural}), and text summarization~\cite{narayan2018don,liu2018controlling}. The networks are designed to capture local contextual information by applying feature extraction over limited sub-regions of the input. With the assumption that words next to each other have relationship and contribute to the way of interpreting each of the words, convolutional neural networks could possibly capture phrasal writing phenomena in a given corpus. The assumption is also used in obtaining well-known word embeddings (\textit{Google word2vec}~\cite{mikolov2013distributed}, \textit{Stanford GloVe}~\cite{pennington2014glove}).

Sentence features $\mathbf{f}_{s_{i}}$  are, then, captured by applying max pooling over the whole sentence. 

\begin{equation}
\mathbf{f}_{s_{i}} = \mbox{max-pooling}_j({\mathbf{f}_{p^{s_i}_{j}}})
\end{equation}
where max-pooling are operated over each dimension of vectors ${\mathbf{f}_{p^{s_i}_j}}$.

With the same max-pooling operation as above, we compute document features as: 

\begin{equation}
\mathbf{f}_d = \mbox{max-pooling}_i(\mathbf{f}_{s_{i}})
\end{equation}

Finally, we apply a multilayer perceptron (MLP) with one hidden and one output layer
\begin{equation}
MLP(\mathbf{x}) = \mbox{sigmoid}(\mathbf{W}_2 \cdot \tanh(\mathbf{W}_1 \cdot \mathbf{x} + \mathbf{b}_1 ) + \mathbf{b}_2)
\end{equation}
 to compute the score of each phrase $p^{s_i}_{j}$ as
\begin{equation}
P\left(p_{j}^{s_i},s_i,d\right) = MLP\left(\left[ \begin{array}{l} 
{\mathbf{f}_{p^{s_i}_{j}}} \\
{\mathbf{f}_{s_{i}}} \\
{\mathbf{f}_d} \\
\end{array}\right]\right)
\end{equation}
where the hidden layer computes the phrase representative features respecting to its local use (limited word window), its enclosing sentence, and its document. The phrase representative features are fed to the output layer to compute phrase score (ranging from 0.0 to 1.0).  

We now describe how to train the phrase scoring model. The main objective is the trained model should assign summary phrases with higher scores than document phrases if the summary belongs to the document and otherwise, assign summary phrases with lower scores than document phrases if the summary does not belong to the document. We denote mean $E$ and standard deviation $std$ of phrase scores $P$ for each document $d$ in the following equations, which we will use to describe our objective as set of constraints, then formulate into loss function to be optimized.

\begin{equation}
E_c=E[P(p_c,c,d)]  \mbox{ where } p_c \in c, c\in d
\end{equation}

\begin{equation}
std_c=std[P(p_c,c,d)]  \mbox{ where } p_c \in c, c\in d
\end{equation}

\begin{equation}
E_s=E[P(p_s,s,d)]  \mbox{ where } p_s \in s, s\in d
\end{equation}

\begin{equation}
std_s=std[P(p_s,s,d)]  \mbox{ where } p_s \in s, s\in d
\end{equation}

\begin{equation}
E_{c,d'}=E[P(p_c,c,d')] \mbox{ where } p_c \in c, c\not\in d' 
\end{equation}
Where $p, c, s, d $ stand for phrase, summary sentence, document sentence, and the whole document respectively. $c\not\in d'$ means $c$ is not a summary of document $d'$.

The main objective is realized by comparing the mean scores of summary phrases and document phrases:

\begin{description}
\item[(o1)] The mean score of summary phrases is higher than the mean score of document phrases: $E_c > E_s$.  
\end{description}

\begin{description}
\item[(o2)] The mean score of summary phrases is lower than document phrases when comparing a summary with a document that the summary does not belong to: $E_{c,d'} < E_{s'}$. This is the negative constraint as opposed to the constraint \textbf{o1}.
\end{description}

The above two constraints are straightforward as the positive and negative factors of the objective. However, the comparison of the mean values does not guarantee to obtain to good scoring model as the score boundaries are not considered yet. 

\begin{description}
\item[(o3)] The maximum score of summary phrases is higher than the maximum score of document phrases. It is expected that there exists concise summary phrases which is typical and representative for the document but could not found in the document. Such summary phrases should get higher scores than document phrases. The estimation $E+std$ is used for representing max instead of hard max, whereby the constraint is realized as $(E_c + std_c)  >  (E_s + std_s)$.
\item[(o4)] The minimum score of summary phrases is higher than the mean score of document phrases. Once again, to emphasize the importance of summary phrases, all summary phrases should get higher score than the average score of document phrases. The estimation $E-std$ is used for representing min instead of hard min, whereby the constraint is realized as $(E_c - std_c)  >  E_s$.
\end{description}

The loss function, hence, is composed from the constraints \textbf{(o1-4)} as follows. 

\begin{equation}
\begin{split}
\mathfrak{L}=\sum_d \max(0,m - (& a_1(E_c - E_s)  \\
+ & a_2(\frac{1}{|\{d'\}|}\sum_{d'\neq d}{E_{s'} - E_{c,d'}}) \\
+ & b_1({(E_c + std_c)  - (E_s + std_s)})  \\
+ & b_2({(E_c - std_c)  - E_s})))  \\
\end{split}
\end{equation}

\subsubsection{Document Vector Composition}
We present our method of composing document vectors from the phrase scoring model.

First, given a document, we obtain its phrase scores and its internal representations: phrase level, sentence level and document level encodings.

Then, we compose the document vector as:
\begin{equation}\label{eq:doc-vec-com}
    \mathbf{g}(d) = \frac{ \sum_{i,j}{P\left(p_{j}^{s_i},s_i,d\right) \times \left[ \mathbf{f}_d ; \mathbf{f}_{s_i} ; \mathbf{f}_{p_j^{s_i}}\right]}}{ \sum_{i,j}{P\left(p_{j}^{s_i},s_i,d\right)} }
\end{equation}

This composition resembles summarization where we weight the document internal representations by its summary. Thus, we call this composition encoded summarization.

\begin{figure}
\centering
\includegraphics[width=0.3\textwidth,trim=0 8cm 26cm 0,clip]{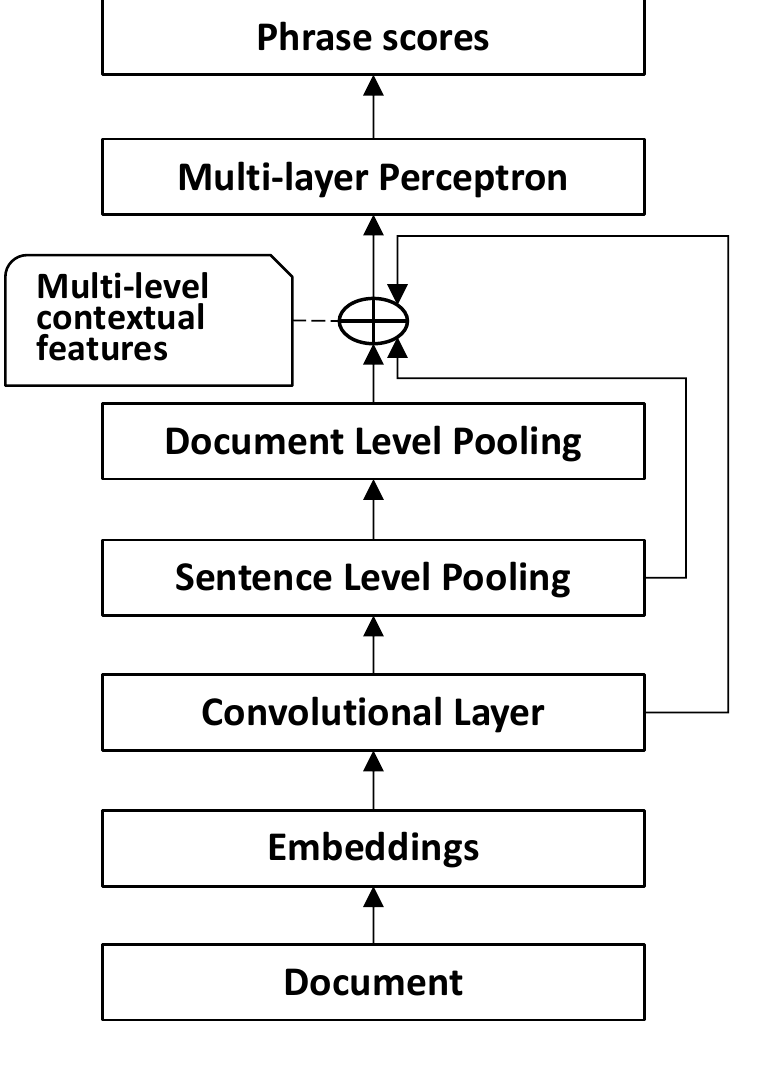}
\caption{Phrase scoring model architecture.}
\end{figure}

\subsubsection{Query-Candidate Relevance Vector}
Each dimension in one document vector is obtained from composition of the same kernel representing one feature.
By comparing each dimension independently, we can estimate the compatibility of a query and a candidate over the dimension. Thus, we compute the query-candidate relevance vector as the element-wise product of query vector and candidate vector.

First, we obtain query vector $\mathbf{g}(q)$ and candidate vector $\mathbf{g}(c)$ using the document vector composition in Equation \ref{eq:doc-vec-com}.
Then, we compute the relevance vector of query $q$ and candidate $c$ by the following element-wise product.
\begin{equation}
    \mathbf{h}(q,c) = \mathbf{g}(q) \odot \mathbf{g}(c)
\end{equation}

\subsubsection{Generating Text Summary}\label{sec:gen-text-sum}

In this phase, we generate a summary given a document by selecting and joining document phrases scored by the phrase scoring model. The process is as follows.
\begin{itemize}
    \item Rank document phrases by they phrasal scores.
    \item Select phrases with scores from high to low.
    \item Join overlapping phrases into a longer phrase.
    \item Stop when the summary length exceeds length-threshold $t$.
\end{itemize}  
The result summary is a list of phrases. The shortest phrases contains $l$ words ($l$ is the window size of the convolutional neural layer). The longest phrases are the sentences themselves.

\subsection{Lexical Matching}\label{sec:lex-matching}
We perform multi-aspect lexical matching where we compare a query case with its candidates via different views. We represent each query as 2 parts: summary and paragraphs, and each candidate as 3 parts: summary, lead sentences (of each paragraph), and paragraphs. This time, most of the candidates do not have a summary, thus, we generate a summary for each candidate by utilizing our encoded summarization method with the summary generation process described in Section~\ref{sec:gen-text-sum}. 

To have more precise comparison between a query versus a candidate, we compute 6 matching options: 
\begin{itemize}
    \item Summary vs. Summary: Matching the summary of the query case with the summary of candidates.
    
    \item Summary vs. Lead-sentences: Matching the summary of the query case with the opening of each paragraph of candidates.
    
    \item Summary vs. Paragraphs: Matching the summary of the query case with the details of candidates.
    
    \item Paragraphs vs. Summary: Matching the details of the query case with the summary of candidates.
    
    \item Paragraphs vs. Lead-sentences: Matching the details of the query case with the opening of each paragraph of candidates.
    
    \item Paragraphs vs. Paragraphs: Matching the details of the query case with candidates.

\end{itemize}

We perform various kinds of text matching including: n-gram, skip-gram, subsequence matching formulas. 

\begin{itemize}
    \item N-gram matching: measuring n-gram overlapping between a query and a candidate case. We employ unigram and bigram models.
    \item Skip-bigram matching: measuring the co-occurrence of all word pairs in their sentence order. This allows the same non-continuous word pairs could be found in both query and candidate.
    \item We also employ the unigram + skip-gram model which balances the unigram matching and skip-gram matching.
    \item Longest common subsequence: measuring the strictly ordered overlapping scattering over the texts. We employ 2 variants: standard version and distance weighted version. The distance weighted version favors subsequences with less distances among words. 
\end{itemize}

For each matching formula, we compute the matching scores by 3 different factors:
\begin{itemize}
    \item Recall: normalized by query, measuring the percentage of the query contents found in the candidate.
    \item Precision: normalized by candidate, measuring the percentage of the candidate contents found in the query.
    \item F-measure: harmony score of the previous two.
    $$f\text{-}measure=\frac{2 \times precision \times recall}{precision+recall}$$
\end{itemize}

For the computation of lexical matching features, we used publicly available ROUGE library\footnote{\url{https://github.com/andersjo/pyrouge}}.

In total, we collect lexical features from 6 matching options, 6 matching formulas and 3 matching factors, which results in lexical feature space with 108 dimensions. 

\subsection{Learning to Rank Candidates}
We formulate the task as ranking problem and devise the learning to ranking method to solve it using lexical matching features from Sections \ref{sec:lex-matching} and query-candidate relevance features from Section \ref{sec:enc-summarization}.

We utilize pair-wise ranking strategy: pairing each supporting case with an irrelevant case from the candidate list. We adopt Linear-SVM as the learning algorithm for solving the optimization problem. The final ranking model should assign high scores to supporting cases and low scores to non-noticed cases. 

After obtaining the scored candidates as a ranked list, we proceed to select top $k$ candidates as the predicted noticed cases. 

\section{Experiments}
\subsection{Experimental Settings}
For the phrase scoring model, we copied the model settings from \cite{vutran2018mirel}, except the loss coefficients, as shown in Table \ref{tbl:sys-params}. The loss coefficients were tuned using random search on COLIEE 2018 dataset. 
For word embeddings, we use the pre-trained GloVe\footnote{\url{https://nlp.stanford.edu/projects/glove/}   Common Crawl (840B tokens, 2.2M vocab, cased, 300d vectors)} which was trained on data from ``Common Crawl"\footnote{\url{http://commoncrawl.org/}}, an open repository of web crawl data. 

The phrase scoring model was trained on COLIEE 2018 dataset, and adopted to generate encoded summarization vectors for case documents, and text summaries for the candidate cases in COLIEE 2019 dataset. For generating the text summaries, the summary length threshold $t$ (Section~\ref{sec:gen-text-sum}) is set to $t = 20\%$ document\text{-}length. The average length of summaries is $\approx{10\%} $ document\text{-}length for COLIEE 2018 dataset (Table~\ref{tbl:data-stat-2018}), and $\approx{9\%}$ document\text{-}length for COLIEE 2019 dataset (Table~\ref{tbl:data-stat-2019}). Thus, with a threshold $t=20\% $ document\text{-}length, we could expect to cover potential information with good recall rate while keeping an acceptable summary length. 

For text pre-processing, we use the default word and sentence tokenization from NLTK~\cite{Loper:2002:NNL:1118108.1118117}\footnote{\url{https://www.nltk.org/} version 3.3.0}. We use ROUGE library with Porter stemmer and stopword removal enabled\footnote{rouge parameters: -c 95 -2 -1 -U -r 1000 -n 2 -w 1.2 -a -d -m -s}.

\begin{table}
\caption{Phrase scoring model parameters.}
\label{tbl:sys-params}
\centering
\begin{tabular}{l|l}
\hline
\hline
\textbf{Parameter} & \textbf{Description} \\
\hline
\hline
Embeddings (vector size $d$) & GloVe \cite{pennington2014glove}
$d=300$  \\
CNN filters $c$ & 300 \\
CNN window size $l$ & 5 \\
MLP hidden size & 300 \\
Optimizer & Adam\cite{duchi2011adaptive} \\
Learning rate & 0.0001 \\
Gradient clipping max norm & 5.0 \\
Loss coefficients
$(a_1,a_2,b_1,b_2)$ & $(1.0,1.7,0.3,0.7)$ \\
Size of negative set $|\{d'\}|$ & 2 \\
loss margin $m$ & 0.5 \\
\hline
\hline

\end{tabular}
\end{table}

We reported our system's validation results  with the following metrics:
\begin{itemize}
    \item MRR: Mean reciprocal rank. This metric measures the rank of the first correct answer or the first (true) supporting case given a query case. In other words, this measures how far users have to read to find a relevant case when looking from the top of the retrieved list.
    \item MAP: Mean average precision. Since we pursue the task with learning to rank method, we use this common ranking evaluation metric. 
    \item Prec, Rec, F1: Precision, Recall, F-1 whose values are averaged by query. This is straightforward as we average the results of all folds in the leave-one-out validation. 
\end{itemize}

For computing Prec, Rec, and F1, we proceed to perform top $k$ selection from the ranked list output by the Linear-SVM Rank model. The selected value of $k$ is 5, the average number of noticed cases in COLIEE 2019 dataset.

For presenting lexical features' impact analysis, we use a coding for representing subsets of the full lexical feature set. The coding is in the form of q-c, where 
\begin{itemize}
    \item q is a subset of query components including summary (s) and paragraphs (p),
    \item c is a subset of candidate components including paragraphs (p), lead sentences (l), and generated summary (e) (described in Section~\ref{sec:gen-text-sum}),
    \item each component in q is compared with each component in c.
\end{itemize}
For example, the lexical method sp-ple (q=sp, c=ple) means we perform all 6 matching options, and the lexical method s-p (q=s, c=p) means we only compare the summary of a query with the paragraphs of a candidate.

\subsection{Validation Results}
The validation results are shown in Table \ref{tbl:loo-val-results}. The best two are ES+sp-ple and ES+sp-pl. ES+sp-ple achieves the best MRR of 0.963 and the best MAP of 0.833.  ES+sp-pl achieved the best Prec of 0.579, Rec of 0.724 and F1 of 0.557.

\begin{table}
    \centering
    \caption{Validation results. The coding for lexical features is in the form of q-c, where q is a subset of query components including summary (s) and paragraphs (p), c is a subset of candidate components including paragraphs (p), lead sentences (l), and generated summary (e) (described in Section~\ref{sec:gen-text-sum}), and each component in q is compared with each component in c. For example, the lexical method sp-ple (q=sp, c=ple) means we perform all 6 matching options, and the lexical method s-p (q=s, c=p) means we only compare the summary of a query with the paragraphs of a candidate. }
    \begin{tabular}{l|c|c|c|c|c}
\textbf{Model}	&	\textbf{MRR}	&	\textbf{MAP}	&	\textbf{Prec}	&	\textbf{Rec}	&	\textbf{F1}	\\ \hline
\textbf{Lexical} & & & & \\
s-p	&	0.843	&	0.690	&	0.484	&	0.620	&	0.470	\\
s-l	&	0.798	&	0.589	&	0.420	&	0.528	&	0.405	\\
s-e	&	0.756	&	0.561	&	0.401	&	0.517	&	0.390	\\
p-p	&	0.845	&	0.680	&	0.476	&	0.601	&	0.461	\\
p-l	&	0.805	&	0.619	&	0.443	&	0.563	&	0.429	\\
p-e	&	0.801	&	0.588	&	0.413	&	0.534	&	0.402	\\[-0.75em] & & & & &\\
sp-p	&	0.871	&	0.712	&	0.490	&	0.635	&	0.480	\\
sp-l	&	0.816	&	0.634	&	0.448	&	0.570	&	0.435	\\
sp-e	&	0.793	&	0.602	&	0.429	&	0.553	&	0.416	\\[-0.75em]
& & & & &\\
sp-pl	&	0.857	&	0.713	&	0.493	&	0.639	&	0.483	\\
sp-pe	&	0.864	&	0.709	&	0.485	&	0.633	&	0.476	\\
sp-ple	&	0.859	&	0.715	&	0.495	&	0.641	&	0.485	\\[-0.75em]
& & & & &\\ \hline
ES	&	0.840	&	0.576	&	0.436	&	0.534	&	0.410	\\[-0.75em]
& & & & &\\
ES + sp-p	&	0.957	&	0.822	&	0.572	&	0.715	&	0.549	\\
ES + sp-pl	&	\textit{0.959}	&	\textit{0.831}	&	\textbf{0.581}	&	\textbf{0.730}	&	\textbf{0.560}	\\
ES + sp-pe	&	0.954	&	0.823	&	0.577	&	0.716	&	0.553	\\
ES + sp-ple	&	\textbf{0.963}	&	\textbf{0.833}	&	\textit{0.579}	&	\textit{0.724}	&	\textit{0.557}	\\ \hline

    \end{tabular}
    \label{tbl:loo-val-results}
\end{table}

The validation results (Table~\ref{tbl:loo-val-results}) of lexical features with various combinations from the 6 matching options in Section~\ref{sec:lex-matching} show that the combination of lexical matching options does have positive effect to improve the performance. sp-p, the combination of the two (s-p and p-p), achieves MAP of 0.712 and F1 of 0.480, which are higher than those of the two. The same effect appears on sp-l, sp-e, sp-pl, sp-ple. Even though with the exception of sp-pe: MAP and F1 are lower than sp-p, sp-ple is the best lexical combination, which wins over other lexical combinations over three of four evaluation metrics with MAP of 0.715, Prec of 0.495, Rec of 0.641, and F1 of 0.485. 

The encoded summarization (ES) approach alone achieves MAP of 0.576 and F1 of 0.410, lower performance than the best lexical combination. The effect is different from the observation in \cite{vu2018coliee} where the performance of encoded summarization is higher than lexical matching approach. Since the encoded summarization model is  trained on only COLIEE 2018 dataset, some summary phenomena in COLIEE 2019 dataset may not be well captured. 

The combination of encoded summarization and lexical features does improve performance significantly. In term of F1 score, the improvement is from +0.068 (ES+sp-pe) to +0.077 (ES+sp-pl). In term of MAP score, the improvement is from +0.108 (ES+sp-pe) to +0.118 (ES+sp-pl and ES+sp-ple). The improvement by the combination shows that, even though the encoded summarization does not perform well alone, it still provides useful information for identifying relevant cases. This effect is also observed in \cite{vu2018coliee}.

\subsection{Submission Results}

We submitted 3 runs to the competition: ES+sp-p, ES+sp-pl, and ES+sp-ple (Table~\ref{tbl:submission-results}). The best test performance is F1 of 0.5764 achieved by two models: ES+sp-pl and ES+sp-ple. Moreover, we also have achieved the best performance compared to other participants of COLIEE 2019 on the test set of the legal case retrieval task (Table~\ref{tbl:competition-results}).

\begin{table}[H]
    \centering
    \caption{Submission results.}
    \begin{tabular}{l|c|c|c}
	\textbf{Model}	&	\textbf{Prec}	&	\textbf{Rec}	&	\textbf{F1}	\\ \hline
	{ES + sp-p}	&	{0.5934}	&	{0.5485}	&	{0.5701}	\\
	{ES + sp-pl}	&	{0.6000}	&	{0.5545}	&	\textbf{0.5764}	\\
	{ES + sp-ple}	&	{0.6000}	&	{0.5545}	&	\textbf{0.5764}	\\ \hline
    \end{tabular}
    \label{tbl:submission-results}
\end{table}

\begin{table}[H]
    \centering
    \caption{Difference in test submission of ES+sp-pl and ES+sp-ple.}
    \begin{tabular}{c|cc|cc}
& 		\textbf{ES + sp-pl}	&		&	\textbf{ES + sp-ple}			\\
\textbf{Query} 	&	\textbf{Candidate}	&	\textbf{Relevant?}	&	\textbf{Candidate}	&	\textbf{Relevant?}	\\ \hline
002	&	043	&	NO	&	181	&	NO	\\
004	&	182	&	NO	&	067	&	NO	\\
007	&	157	&	NO	&	023	&	NO	\\
009	&	003	&	NO	&	107	&	NO	\\
\textbf{010}	&	\textbf{002}	&	\textbf{NO}	&	\textbf{164}	&	\textbf{YES}	\\
013	&	187	&	NO	&	132	&	NO	\\
014	&	022	&	NO	&	020	&	NO	\\
016	&	168	&	NO	&	115	&	NO	\\
017	&	126	&	NO	&	073	&	NO	\\
\textbf{028}	&	\textbf{113}	&	\textbf{YES}	&	\textbf{039}	&	\textbf{NO}	\\
\textbf{030}	&	\textbf{116}	&	\textbf{YES}	&	\textbf{044}	&	\textbf{YES}	\\
033	&	126	&	NO	&	028	&	NO	\\
\textbf{035}	&	\textbf{001}	&	\textbf{YES}	&	\textbf{066}	&	\textbf{NO}	\\
037	&	116	&	NO	&	188	&	NO	\\
039	&	070	&	NO	&	095	&	NO	\\
042	&	001	&	NO	&	032	&	NO	\\
\textbf{048}	&	\textbf{163}	&	\textbf{NO}	&	\textbf{121}	&	\textbf{YES}	\\
051	&	090	&	NO	&	175	&	NO	\\
056	&	079	&	NO	&	101	&	NO	\\
    \end{tabular}
    \label{tbl:cmp-test-outputs}
\end{table}

Even though ES+sp-pl and ES+sp-ple share the same test performance as shown in Table~\ref{tbl:submission-results}, the test submissions are different. Over 61 test queries, 19 queries are submitted differently for each of the two models. The noticeable differences are of queries 010, 028, 030, 035, and 048, where the performance changes. In the cases of queries 010 and 048, ES+sp-ple has better outputs, discards irrelevant candidates and selects relevant ones. In the cases of queries 028 and 035, ES+sp-ple has worse outputs, discards relevant candidates and selects irrelevant ones. In the last case of query 030, ES+sp-pl and ES+sp-ple choose different but all relevant candidates. The two models differ in the addition of comparing a query's summary and paragraphs to a candidate's generated summary. 

\begin{table}
    \centering
    \caption{Participants' results. We submitted 3 runs to the competition: ES+sp-p, ES+sp-pl, and ES+sp-ple, corresponding to JNLP.task\_1.p, `JNLP.task\_1.pl, and JNLP.task\_1.ple, respectively. We achieved the best performance of 0.5764 F1 score on the test set of the legal case retrieval task. }
    \begin{tabular}{l|l|c|c|c}
\textbf{Team}	&	\textbf{Run name}	&	\textbf{Prec}	&	\textbf{Rec}	&	\textbf{F1}	\\ \hline
CACJ	&	submit\_task1\_CACJ01	&	0.2119	&	0.5848	&	0.3110	\\
CLArg	&	CLarg	&	0.9266	&	0.3061	&	0.4601	\\
HUKB	&	task1.HUKB	&	0.7021	&	0.4000	&	0.5097	\\
IITP	&	task1.IITPBM25	&	0.6256	&	0.3848	&	0.4765	\\
IITP	&	task1.IITPd2v	&	0.4653	&	0.3455	&	0.3965	\\
IITP	&	task1.IITPdocBM	&	0.6368	&	0.3879	&	0.4821	\\
ILPS	&	BERT\_Score\_0.946	&	0.6810	&	0.4333	&	0.5296	\\
ILPS	&	BERT\_Score\_0.96	&	0.8188	&	0.3424	&	0.4829	\\
ILPS	&	BM25\_Rank\_6	&	0.4672	&	0.5182	&	0.4914	\\
\textit{JNLP}	&	\textit{JNLP.task\_1.p}	&	{0.5934}	&	{0.5485}	&	\textit{0.5701}	\\
\textbf{JNLP}	&	\textbf{JNLP.task\_1.pl}	&	{0.6000}	&	{0.5545}	&	\textbf{0.5764}	\\
\textbf{JNLP}	&	\textbf{JNLP.task\_1.ple}	&	{0.6000}	&	{0.5545}	&	\textbf{0.5764}	\\
UA	&	UA\_0.52	&	0.3513	&	0.3364	&	0.3437	\\
UA	&	UA\_0.54	&	0.3639	&	0.3242	&	0.3429	\\
UA	&	UA\_0.57	&	0.3560	&	0.3333	&	0.3443	\\ \hline
    \end{tabular}
    \label{tbl:competition-results}
\end{table}

\section{Conclusion}
We have presented our method for legal case retrieval task in Competition on Legal Information Extraction/Entailment, 2019. We have applied the approach of combining lexical features and latent features embedding summary properties which we call encoded summarization. The combination of encoded summarization and lexical features does improve performance significantly. Besides, the inclusion of lexical matching with lead sentences of each paragraph and text summaries generated via the phrase scoring model shows positive effect in improving retrieval performance. For validation results, we achieved best MAP score of 0.833 and second best F1 score of 0.557 (the best is 0.560) with combination of encoded summarization with all of the described lexical features including comparing a query's summary and paragraphs to a candidate's paragraphs, lead sentences and summary generated via the phrase scoring model. We have showed that the phrase scoring model trained from COLIEE 2018 dataset can provide useful features for representing documents in COLIEE 2019 dataset. 

There are several directions for improving the performance of legal case retrieval systems. 
One is that we can use the documents having a summary in COLIEE 2019 dataset for fine-tuning the phrase scoring model. Besides, the lexical matching has not yet considered the statistical information of terms in the corpus, which can be modeled by term frequency-inverse document frequency for example. Including such information may improve the matching by recognizing the statistically typical words for each document. 

\begin{acks}
This work was supported by JST CREST Grant Number JPMJCR1513, Japan.
\end{acks}

\bibliographystyle{ACM-Reference-Format}
\bibliography{main}
\end{document}